\documentclass[letterpaper, 10 pt, conference]{IEEEtran} 

\usepackage{graphics} 
\usepackage{epsfig} 
\usepackage{amsmath} 
\usepackage{amssymb}  

\usepackage[utf8]{inputenc} 
\usepackage[T1]{fontenc}    

\usepackage[dvipsnames]{xcolor}
\usepackage{booktabs}       
\usepackage{amsfonts}       
\usepackage{nicefrac}       
\usepackage{microtype}      

\usepackage{booktabs} 
\usepackage{multirow}
\usepackage{pifont}
\usepackage{setspace}
\usepackage{comment}
\usepackage{algorithm}
\usepackage[noend]{algpseudocode}

\usepackage{verbatim}
\usepackage{caption}
\usepackage{wrapfig}
\usepackage{svg}
\usepackage{mathabx}
\usepackage{subcaption}
\usepackage{enumitem}
\usepackage{color,soul}
\usepackage{xfakebold}

\usepackage[noadjust]{cite}
\usepackage[english]{babel}
\usepackage{amsthm}
\theoremstyle{definition}

\theoremstyle{remark}

\setul{0.5ex}{0.3ex}
\definecolor{grey}{rgb}{.80,.80,0.80}
\setulcolor{grey}

\usepackage{hyperref}
\hypersetup{
    colorlinks=true,
    linkcolor=orange,
    filecolor=magenta,      
    urlcolor=orange,
    citecolor=orange,
}

\IEEEoverridecommandlockouts                              

\usepackage{amsfonts}               
\usepackage{amsmath}                
\usepackage{booktabs}               
\usepackage[T1]{fontenc}            
\usepackage{graphicx}               
\usepackage{hyperref}               
\usepackage{inconsolata}            
\usepackage[numbers,sort&compress]{natbib}        
\usepackage{nicefrac}               
\usepackage{xcolor}                 
\usepackage{multicol}

\usepackage[utf8]{inputenc} 
\usepackage[T1]{font enc}    
\usepackage{hyperref}       
\usepackage{url}            
\usepackage{booktabs}       
\usepackage{amsfonts}       
\usepackage{nicefrac}       
\usepackage{microtype}      
\usepackage{xcolor}         

\hypersetup{
    colorlinks=true,
    linkcolor=orange,
    filecolor=magenta,      
    urlcolor=orange,
    citecolor=orange,
}

\usepackage{titlesec}
\usepackage{xspace}
\usepackage{mathtools}
\usepackage{bbm}
\usepackage{enumitem}
\usepackage{caption}
\usepackage{subcaption}
\usepackage{pdflscape}
\usepackage{xcolor}

\setcitestyle{numbers,square,comma}
\usepackage{amsthm}
\usepackage{amsmath}

\usepackage[capitalise, nameinlink]{cleveref}

\newcommand{\fbseries}{\unskip\setBold\aftergroup\unsetBold\aftergroup\ignorespaces}
\makeatletter
\newcommand{\setBoldness}[1]{\def\fake@bold{#1}}
\newcommand{\ssymbol}[1]{^{\@fnsymbol{#1}}}
\makeatother

\title{
\Large \bf
How to Train Your Robots? \\
The Impact of Demonstration Modality on Imitation Learning
}

\author{
Haozhuo Li$\ssymbol{3}$, Yuchen Cui$\ssymbol{4}$, Dorsa Sadigh$\ssymbol{3}$
\thanks{ 
\hspace{-0.37cm} $\ssymbol{3}$ Computer Science Department, Stanford University. \newline
 $\ssymbol{4}$ Computer Science Department, University of \noindent California, Los Angeles. \newline
\noindent Corresponding Email: {{\tt\footnotesize yuchencui@cs.ucla.edu}}
}
}

\begin{document}

\maketitle
\thispagestyle{empty}
\pagestyle{empty}

\begin{abstract}
Imitation learning is a promising approach for learning robot policies with user-provided data. The way demonstrations are provided, i.e., demonstration modality, influences the quality of the data. 
While existing research shows that kinesthetic teaching (physically guiding the robot) is preferred by users for the intuitiveness and ease of use, the majority of existing manipulation datasets were collected through teleoperation via a VR controller or spacemouse.
In this work, we investigate how different demonstration modalities impact downstream learning performance as well as user experience. 
Specifically, we compare low-cost demonstration modalities including kinesthetic teaching, teleoperation with a VR controller, and teleoperation with a spacemouse controller.
We experiment with three table-top manipulation tasks with different motion constraints. 
We evaluate and compare imitation learning performance using data from different demonstration modalities, and collected subjective feedback on user experience.
Our results show that kinesthetic teaching is rated the most intuitive for controlling the robot and provides cleanest data for best downstream learning performance. 
However, it is not preferred as the way for large-scale data collection due to the physical load. Based on such insight, we propose a simple data collection scheme that relies on a small number of kinesthetic demonstrations mixed with data collected through teleoperation to achieve the best overall learning performance while maintaining low data-collection effort.
\end{abstract}

\section{Introduction}
\label{sec:introduction}

Imitation learning has shown to be a promising approach for scaling up learning end-to-end robot policies for diverse tasks through collecting ad-hoc demonstrations from end-users. 
However, the performance of imitation-based policies depends in large part on the \emph{quality} of the demonstrated data~\cite{mandlekar2021robomimic,belkhale2024data,hejnaremix}, which can be influenced by a number of design factors during the data collection process.
In particular, the \emph{way} of providing demonstrations, which we refer to as \emph{demonstration modality}, is an important factor that can affect data quality by influencing how users control the robot~\cite{cui2021understanding}. 
\begin{figure}
    \centering
    \includegraphics[width=0.9\linewidth]{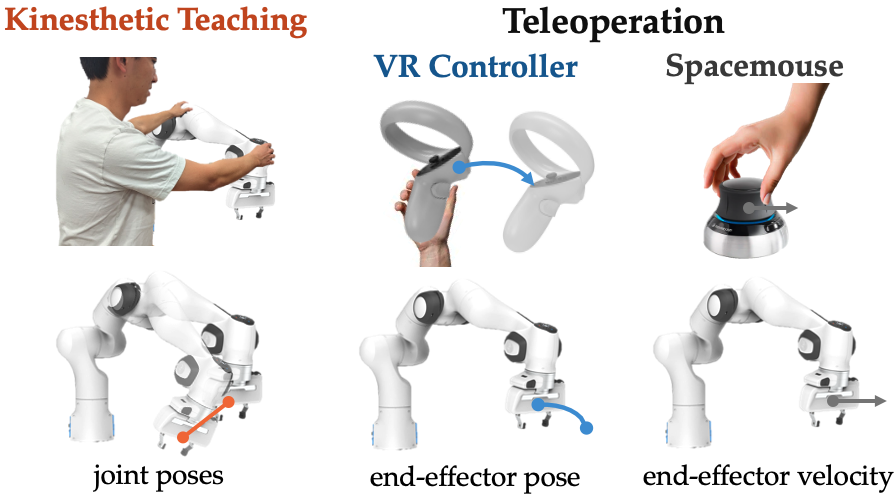}
    \caption{\textbf{Demonstration modalities under study.} Kinesthetic teaching controls precise joint poses; and teleoperation controls the delta pose of the robot's end-effector: VR provides a direct spatial mapping of the trajectory, while spacemouse allows the user to command velocity through buttons.}
     \vspace{-0.4cm}
    \label{fig:modalities}
\end{figure}

\begin{figure}
    \centering
    \includegraphics[width=0.8\linewidth]{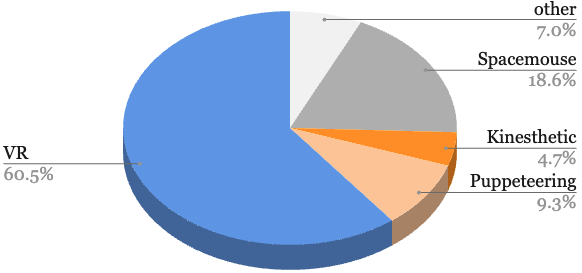}
    \caption{\textbf{Popular demonstration modalities.} Composition of human demonstration modalities present in the OpenXE dataset~\cite{open_x_embodiment_rt_x_2023}.}
    \label{fig:modality_percentage}
    \vspace{-0.35cm}
\end{figure}

There are various ways to provide a demonstration, including \emph{physically guiding} the robot through kinesthetic teaching~\cite{billard2008survey,akgun2012trajectories,sharma2018multiple} or puppeteering via a leader robot~\cite{zhao2023learning,yang2024ace}, and \emph{teleoperating} the robot with different control devices such as 3D spacemouse controllers~\cite{zhu2022bottom,liu2023robot} or virtual reality (VR) interfaces~\cite{rakita2018autonomous,stotko2019vr,mandlekar2018roboturk,mandlekar2021robomimic}. These modalities, shown in~\cref{fig:modality_percentage}, often compose the majority of existing robot datasets such as the OpenXE dataset~\cite{open_x_embodiment_rt_x_2023}.

While there are many different ways of providing robot demonstrations, teleoperation with a VR controller or a spacemouse is currently the dominant data collection modality for training visuomotor policies.
This is despite the fact that \emph{kinesthetic teaching} is shown to be more intuitive for people to use compared to teleoperation methods~\cite{duan2023ar2,jiang2024comprehensive}.
\emph{Kinesthetic teaching} refers to the demonstrator physically moving a passive robot arm to complete a given task. The demonstrator can control the precise position of each joint of the robotic arm. 
Since the robot's controller is not engaged during data collection, replaying the recorded trajectory is necessary to determine the actions needed to replicate each pose. Replaying also yields clean visual observations unimpeded by the demonstrator’s hand. While kinesthetic teaching is often more intuitive, it also comes with additional costs in time and physical effort.
By contrast, teleoperation modalities (e.g., VR or spacemouse) allow the demonstrator to directly control the end-effector’s motion. 
However, these two modalities also have some nuanced differences.
With \emph{VR teleoperation}, the demonstrator only needs to have a rough estimate of how their motion scales to the robot's motion and their hand trajectory can then directly translate to robot trajectory. The demonstrator hence has a direct spatial control of the motion. 
With \emph{spacemouse} or joystick-style controllers, the demonstrator directly controls the magnitude and direction of the end-effector’s  motion using combinations of button presses. 

Despite the wide range of demonstration modalities, their relative effects on \textit{policy performance}, \textit{data quality}, and \textit{user experience} remain insufficiently explored. In this work, we investigate how these distinct demonstration modalities influence imitation learning for manipulation tasks through these three lenses.
We focus on studying common \emph{low-cost} data collection modalities~(shown in~\cref{fig:modalities}) for robotic arms used by practitioners, including kinesthetic teaching, teleoperation through VR controller, and teleoperation through spacemouse.

We evaluate the effect of these three demonstration modalities via different manipulation tasks with varying motion constraints as our testbed. 
Additionally, we propose a data collection paradigm that leverages the benefits of these different modalities.
We first collect demonstrations from a single expert across three different modalities for all the tasks and train diffusion-based imitation policies~\cite{chi2023diffusionpolicy}.
We then conduct a user study where participants provide demonstrations for the manipulation tasks using all three modalities and compare the modalities in terms of subjective and objective metrics.
Our key findings are:
\begin{itemize}
[itemsep=0.2em,nolistsep,labelindent=0.5em,labelsep=0.15cm,leftmargin=*]
    \item Kinesthetic teaching produces data that leads to higher policy performance and is preferred by users for the ease of use when controlling robots.
    \item However, kinesthetic teaching is not preferred by users for large-scale data collection due to its physical demand and added time to replay.
    \item Data from kinesthetic teaching exhibits high action consistency, while teleoperation via VR or spacemouse offer higher state diversity, where both metrics of action consistency and state diversity correlate with high quality data.
\end{itemize}

Based on these insights, we propose a \emph{hybrid data collection scheme} that combines a small amount of data from kinesthetic teaching with additional data from VR teleoperation. This simple approach results in an average of 20\% higher performance than using data from individual modalities alone, while maintaining a low physical burden on the demonstrator.

\section{Related Work}
\label{sec:related-work}

Our work is broadly related to research that designs control interfaces and data collection methods for imitation learning, as well as work in human-robot interaction that aims to understand interface design implications on user experience. 
We also borrow tools from recent research on data quality in imitation learning to analyze data from different modalities.

\medskip

\noindent  \textbf{Demonstration modalities for imitation learning.}
Demonstration modalities are generally categorized into \emph{teleoperation} versus \emph{shadowing}, depending on whether there is a direct recording mapping of actions~\cite{argall2009survey,ravichandar2020recent}. 
Teleoperation via a VR controller or spacemouse has become the dominant modality in the era of end-to-end visuomotor policy learning due to the accessibility of these low-cost devices. 
Spacemouse is similar to traditional joystick controllers where the user is effectively pressing buttons to control each axis of motion, but it requires only one hand to control all 6 degrees of freedom~\cite{zhu2022bottom,liu2023robot}. 
VR controller tracks hand motion with an IMU sensor and therefore can be used to command the delta motion of the robot arm~\cite{stotko2019vr,mandlekar2018roboturk,mandlekar2021robomimic}.
Kinesthetic teaching involves the demonstrator physically moving the robot and requires replaying to recover the \emph{commanded} actions, especially in the presence of contact forces; and hence it is a \emph{shadowing} approach that does not directly record the robot's actions.

The main benefit of kinesthetic teaching over teleoperation is that the demonstrator can physically feel the joint limits and contact forces through the robotic arm.
In an effort to reproduce similar feedback in teleoperation methods, an active body of research develops specialized devices for effective teleoperation such as haptic controllers~\cite{el2020review}. 
Puppeteering, where the demonstrator kinesthetically moves one robot to control an identical twin, is a special case along this direction where a passive robot is used as the controller to provide feedback of joint limits mechanically~\cite{zhao2023learning,yang2024ace}.
Puppeteering shares many characteristics of kinesthetic teaching but bypasses the challenges of replay by recording the commanded poses on the leader robot. 
Pushing the idea of \emph{wearing the robot}, 
a line of research proposes to make the human demonstrator hold a robot gripper to provide data and then use the in-hand camera observation to align states across data collection and deployment~\cite{young2021visual,chi2024universal}. 
Recent research has also made significant progress towards vision-based teleoperation systems, especially mapping human hand motions to robot actions~\cite{duan2023ar2,qin2023anyteleop,bunny-visionpro}.

\smallskip

\noindent  \textbf{User experience of demonstration modality.}
The influence of demonstration interface design has been studied in a number of prior works. For example,
\citet{akgun2012trajectories} show that different types of kinesthetic teaching (keyframe versus trajectory) can have different implications on data quality and user experience.
\citet{wrede2013user} and
\citet{sakr2020training} study how to train or aid non-expert users to better program the robot with kinesthetic teaching.
\citet{rakita2018autonomous} studies user experience of a VR-based remote teleoperation system.
\citet{duan2023ar2} demonstrate an augmented reality-based modality with matching performance with kinesthetic teaching in simple pick-and-place tasks.
Closely related to our work, \citet{jiang2024comprehensive} studies the effect of data collection modalities on user experience and performance in terms of how successful they were able to perform assigned tasks. 

Our contributions differ from prior work in the following aspects: 1) while in prior work performance is evaluated as how successful the user is able to provide demonstrations through a modality, we close the evaluation loop by training downstream imitation policies with the collected data and evaluating the learning performance; and 2) unlike prior work that heavily focuses on the experience of novice users, we are interested in the experience of expert users of robots, especially those who want to collect large-scale data for training effective policies.
Results from our user study align with findings in prior work that most users find kinesthetic teaching more intuitive than teleoperation methods and further confirm that kinesthetic data provides the best performing models except when contact forces are present.
Our user study suggests that most users would choose teleoperation over kinesthetic teaching for large-scale data collection, despite the fact that they rate kinesthetic teaching more favorable in terms of subjective measures. We further propose a simple yet effective data collection scheme that mixes data from multiple demonstration modalities to balance the trade-off between performance and cost.

\medskip

\noindent \textbf{Data quality.}
Scaling real-world robot data for imitation learning is expensive. Therefore, an increasing line of research studies how to formalize, evaluate, and control data quality~\cite{cui2021understanding,gandhi2022eliciting,belkhale2024data,hejnaremix,gao2024}. Specifically, \citet{hejnaremix} uses robust optimization to find the optimal weights for mixing existing datasets in large-scale imitation learning. Other works guide data collection by querying for trajectories with more consistent actions~\cite{gandhi2022eliciting} or trajectories that are guided by compositional generalization capabiliteis of imitation policies~\cite{gao2024}.
By contrast, our work focuses on studying the effect of demonstration modality on collecting task-specific data for imitation learning. Our findings are complementary to prior work and can add additional guidance for future large-scale data collection design. 

\section{Preliminaries}
\label{sec:prelim}

\noindent \textbf{Problem formulation.} Imitation learning assumes access to a demonstration dataset $\mathcal{D}= \{\tau_1,\dots,\tau_N\}$ of $N$ demonstrations. 
Each demonstration $\tau_i$ consists of a sequence of continuous state-action pairs of length $T_i$, $\tau_i = \{(s_1,a_1),\dots,(s_{T_i},a_{T_i})\}$, with states $s \in \mathcal{S}$ and actions $a \in \mathcal{A}$. 
Demonstrations are generated by expert policy $\pi_E(s)$ under environment dynamics $\rho(s'|s,a)$. The objective of imitation learning is to learn a policy $\pi_\theta: \mathcal{S} \to \mathcal{A}$ that maps states to actions similar to $\pi_E$. 

In practice, human demonstrators do not have direct access to the robot's action space and therefore human's internal policy $\pi_\mathcal{H}$ has to go through control modality modifier $\phi_x$. 
The observed dataset hence contains actions from a modality-specific policy:
\begin{equation}
    (s,a) \sim \phi_x(\pi_\mathcal{H}(s),\rho)
\end{equation}
$\phi_x$ captures the biases introduced by different demonstration modality design. For example, joystick-style controllers allow users to easily move in straight lines.

\medskip

\noindent \textbf{Policy model.} We follow the work of \citet{chi2023diffusionpolicy} to formulate the behavioral cloning policies as denoising diffusion probabilistic models that optimize the following loss over $\mathcal{D}_N$:
\begin{equation}
    \mathcal{L(\theta)} = 
    {\text{\emph{MSE}}(\epsilon^k,\epsilon_\theta(s_i, a^0_i+\epsilon^k, k)),}
    \label{eq:bc_loss}
\end{equation}
where $a_i^{K}$ is sampled from Gaussian noise and the process takes $K$ steps of denoising iteration to $a_i^{0}$:
\begin{equation}
    a_i^{k-1} = \alpha (a_i^k - \gamma\epsilon_\theta(s_i,a_i^k,k) + \mathcal{N}(0,\sigma^2I))
\end{equation}
The state $s$ consists of RGB observations that are encoded with ResNet~\cite{he2016deep} and then concatenated with proprioception data. 
We use high-dimensional temporal action sequence as the output space for the diffusion policy models, where one action consists of the desired delta pose of end-effector and a binary indicator for controlling the gripper.

\begin{figure}
    \centering
    \includegraphics[width=0.88\linewidth]{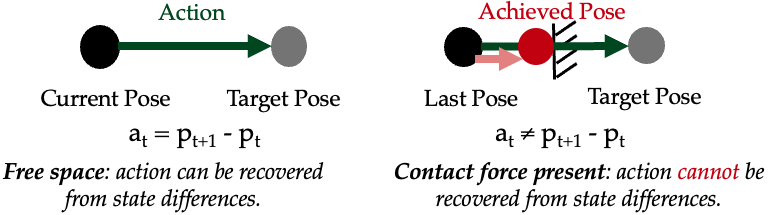}
    \caption{\textbf{Action discrepancy.} Replaying recorded end-effector pose may not recover the desired action, especially when contact force is present.}
    \label{fig:action-diff}
    \vspace{-0.45cm}
\end{figure}

\medskip

\noindent \textbf{Data quality.} We follow the work of \citet{belkhale2024data} to approximate quality of a dataset through measuring action consistency and state diversity in the dataset.  
Instead of clustering states with a fixed distance, we approximate action variance (opposite of action consistency) among $K$ nearest-neighbor states in proprioceptive state space:
\begin{equation}
\label{eq:av}
    \text{ActionVariance}(\mathcal{D}) = \frac{1}{|\mathcal{D}|}\sum_{(s,a)\in\mathcal{D}} (a - \frac{1}{K}\sum_{(\hat{s},\hat{a}) \in \text{\scriptsize{\emph{NN}}}(s, \mathcal{D}, K)}\hat{a}) ^ 2
\end{equation}
We estimate state diversity with the same method by computing state variance instead.
While these metrics do not take into account visual states, the initial state distribution is the same across modalities, therefore the same estimation bias equally exists in all datasets under comparison.

\section{Experimental Design}
\label{sec:method}

In this section, we outline our experimental setup and design choices for evaluating the impact of demonstration modalities. We begin by presenting our environment setup and task design, followed by a detailed description of the data collection pipeline for each modality. We then provide an overview of the policy training process and describe the user study that assesses user experiences across the three demonstration modalities.

\medskip
\noindent \textbf{Setup.} We use a 7-DoF Franka Emika Panda robotic arm as the hardware. We use the Cartesian impedance controller in Polymetis to control desired end-effector pose of the arm. 
We process the user input from different demonstration modalities to compute delta end-effector pose as action commands, such that the collected data from different modalities shares the same state and action representation.

\begin{figure}
    \centering
    \includegraphics[width=\linewidth]{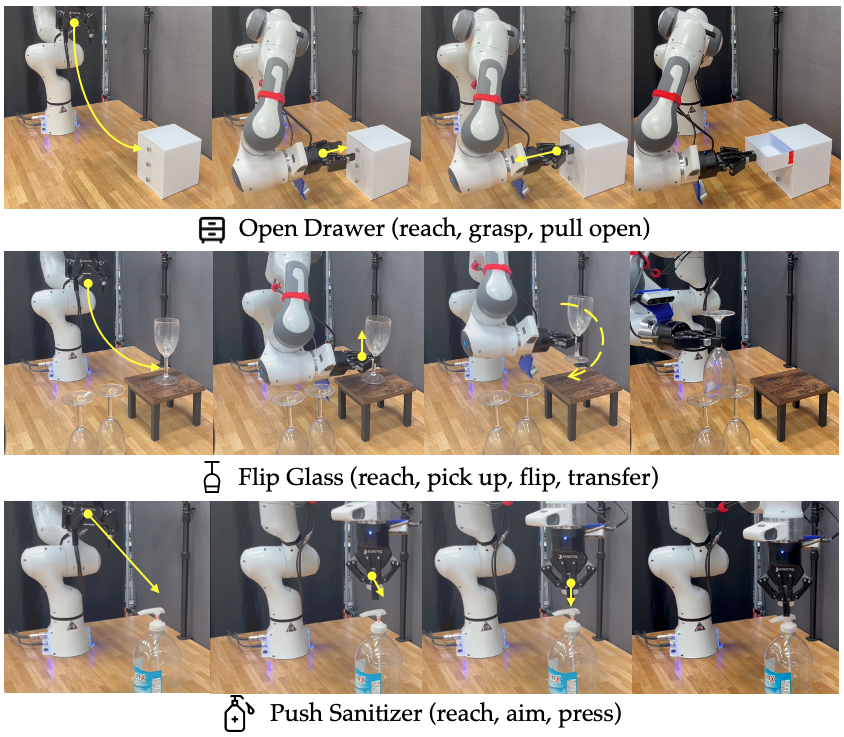}
    \caption{\textbf{Tasks with varying motion constraints.} The three selected tasks each has a different type of motion constraint (e.g. constrained linear, large rotation, and exerting contact force) to represent a broad range of tasks.}
    \vspace{-0.45cm}
    \label{fig:tasks}
\end{figure}

\medskip
\noindent \textbf{Tasks.} We select three manipulation tasks with different types of motion constraints as our testbed, including one requiring high contact force. \cref{fig:tasks} shows the three tasks we study. \emph{Open Drawer} consists of a free-space reaching motion, a precise alignment motion for grasping the knob, and a final constrained linear motion to pull out the drawer. 
\emph{Flip Glass} consists of a free-space reaching motion followed by a pick up motion, a 180-degree rotation to turn the glass upside down, and finally a transfer to the desired placement location. To perform this task, the robot arm needs to operate near its joint limits. 
\emph{Push Sanitizer} involves the robot aiming at the center of the sanitizer top and pressing down to dispense sanitizer. The pushing motion requires pressing force towards the sanitizer. 

\medskip
\noindent \textbf{Data collection interfaces.} For kinesthetic teaching, we first record the end-effector poses and gripper actions $\{(p^\text{demo}_t,g^\text{demo}_t)\}$ in gravity compensated mode as the demonstrator physically moves the robot arm to complete the task. We record at the same frequency as the active robot controller uses. We then reset the task and replay the demonstration using the timestamped end-effector poses as targets and compute the delta pose from the robot's current pose as the action command:
\begin{equation}
    a^0_t = p^\text{demo}_t - p^0_{t} 
\end{equation}
where $p^i_t$ is the pose in $i$-th iteration of replay at timestamp $t$. 
Corresponding states and commanded actions are recorded as the dataset. 
However, the computed actions may not recover the demonstrated trajectory, especially in the presence of contact forces. 
As illustrated in \cref{fig:action-diff}, while the action is equivalent to the delta pose for free space motions, it is nontrivial to recover the action when contact forces are present and the robot may fail to reach the commanded pose.
The principled way to resolve this issue is to record the end-effector force profile during demonstration and replay accordingly~\cite{kormushev2011imitation,montebelli2015handing}. However, we do not have access to a force sensor and can only account for such control error in hindsight using a trick to compensate actions with errors in the previous iteration of replay.
In our data collection, if the replay fails to complete the task in the first try, we will replay with additional error term added to the action to accommodate force-induced error:
\begin{equation}
    a^1_t = p^\text{demo}_t - p^1_{t+1} + \lambda (p^\text{demo}_{t+1}-p^0_{t+1}) 
\end{equation}
This heuristic results in doubling the replay time for tasks that require strong contact forces such as Push Sanitizer.

For VR teleoperation, we use Oculus Quest's controller and access the controller's pose change for controlling the robot~\cite{OrbikEbert2021OculusReader}. 
Pressing a button on the VR controller activates teleoperation, recording the controller’s pose as a reference. The robot's end-effector moves based on the controller’s motion relative to this reference. Releasing the button stops the robot, allowing the user to reset the reference pose for full rotational control.
With spacemouse teleoperation, the input is directly mapped to robot end-effector delta pose with a scaling factor. 

\medskip
\noindent \textbf{Training data.} To control confounding factors, including demonstrator style and experience, we use data from a single demonstrator who is similarly experienced in using all modalities for training policies. 
The demonstrator was previously unexposed to any modality and practiced using each equally prior to data collection.  
Upon data collection, the demonstrator provided a total of 100 trajectories per modality for each task, switching modality every 10 trajectories and marking initial states to ensure the data collected are under conditions as similar as possible.

\medskip
\noindent \textbf{Policy learning.} Data quality is a function of the downstream learning algorithm~\cite{belkhale2024data}, therefore we fix it to be diffusion policy~\cite{chi2023diffusionpolicy}. 
Specifically, the model takes two RGB observations (one wrist camera view and one external camera view) and the robot's proprioception state as input and outputs a sequence of actions. We train the model with supervised loss as sepcified in \cref{eq:bc_loss}. The model is trained to predict 16 consecutive actions, but we only execute the first 8 per inference during evaluation.
We train each model for 800K gradient steps and probe the performance of the last 5 checkpoints (10K steps apart) with a set of in-distribution tests, and select the model with highest performance for full evaluation.
The full evaluation includes half in-distribution and half out-of-distribution tests. We report the success rate out of 20 total trials. 

\medskip
\noindent \textbf{User study.} We conducted a user study to evaluate how user experience is influenced by different demonstration modalities:
\begin{itemize}
[itemsep=0.2em,nolistsep,labelindent=0.5em,labelsep=0.15cm,leftmargin=*]
    \item \textbf{Participants.} We recruited 12 university students (4 female, 8 male, aged 19–29) majoring in computer science or engineering. 6 participants identified as \emph{Expert} in training robot policies, while the rest had little to no experience.
    
    \item \textbf{Independent Variables.}  Participants engaged with 3 demonstration modalities: kinesthetic, VR, and spacemouse. Our results shows no statistically significant differences between VR and spacemouse. Hence we classify VR and spacemouse as teleoperation methods in our subsequent analysis.

    \item \textbf{Dependent Variables.} We collected both objective and subjective measures from the user study. We recorded the time participants spent on the practice task before they feel confident to collect successful demonstrations. 
    We also collected subjective feedback through likert-scale questions in the NASA-TLX survey~\cite{nasa-tlx} for each modality and a comprehensive questionnaire comparing all modalities.

    \item \textbf{Hypotheses.} We have the the following hypotheses: \\
    \textbf{H1.} (\textit{intuitiveness}) kinesthetic teaching requires shorter practice time than teleoperation methods; \\
    \textbf{H2.} (\textit{perceived effectiveness}) kinesthetic teaching is preferred by users for controlling robot while teleoperation methods are preferred for large-scale data collection.

     \item \textbf{Procedure.} Participants first practiced a test task under each modality until they were confident to perform study tasks. Then they are asked to provide 5 successful demonstrations per modality across two assigned tasks—Open Drawer or Push Sanitizer, and Flip Glass. The order of modalities is randomized within a task. After each modality, participants filled out a NASA-TLX survey. A final comprehensive questionnaire was used to compare modalities, asking about preferences for robot control and large-scale data collection.
\end{itemize}

\section{Results}
\label{sec:data_analysis}

In this section, we first present the impact of different modalities on policy learning, then discuss the results of user study, and provide an analysis of the collected data using data quality metrics. 
Finally, we present and discuss a novel data collection scheme based on the insights from our data analysis.

\medskip

\begin{figure}
    \centering
    \includegraphics[width=0.85\linewidth]{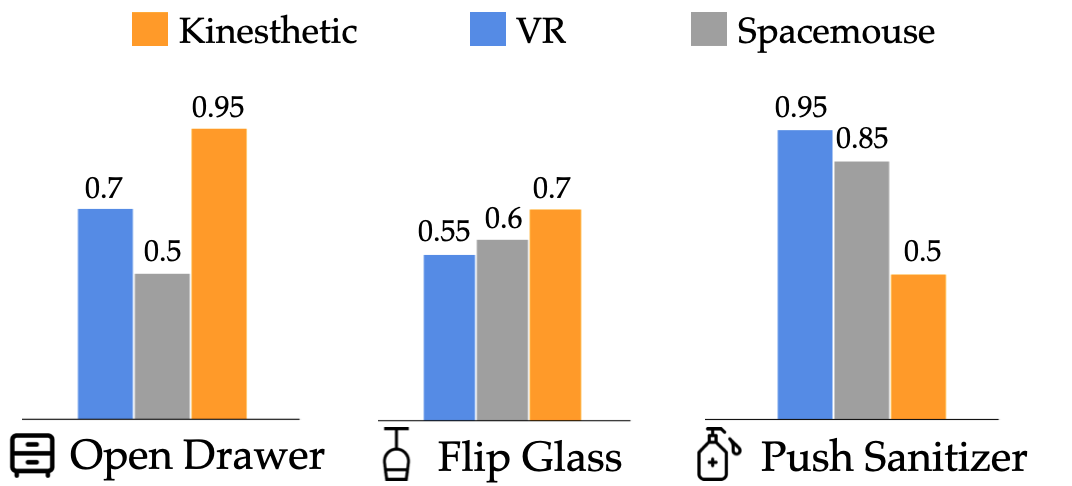}
    \caption{\textbf{Policy performance.} Success rates of policies learned for each task under different demonstration modality. Kinesthetic data leads to best-performing models in Open Drawer and Flip Glass but underperforms in Push Sanitizer where strong contact force is required to complete the task.}
    \label{fig:success_rate}
    \vspace{-0.2cm}
\end{figure}

\begin{figure}
    \centering
    \includegraphics[width=0.8\linewidth]{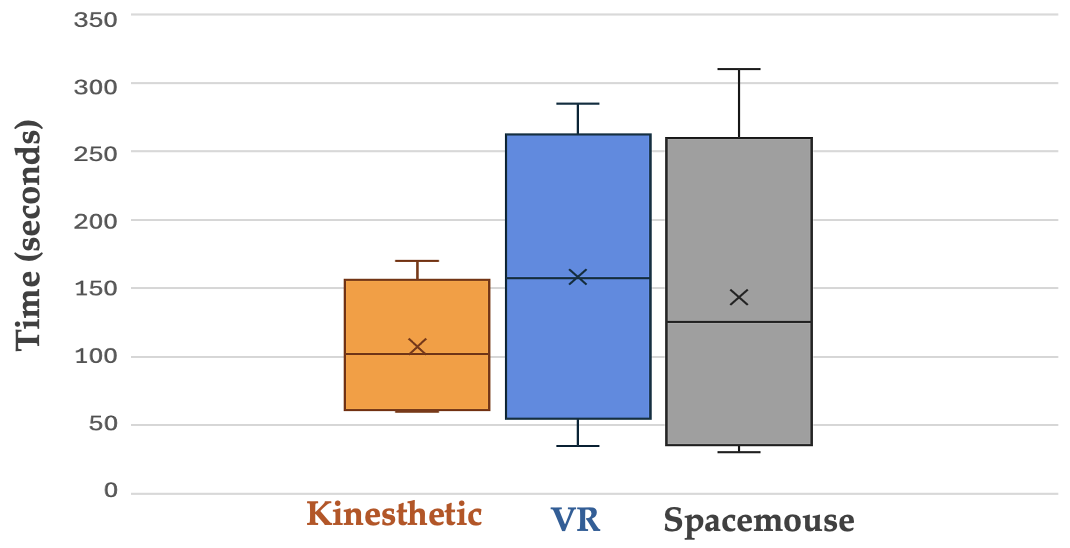}
    \caption{\textbf{Practice time.} Candle chart for the time participants spent on the practice task under each modality until confident. Users spent less time practicing kinesthetic teaching than teleoperation methods (p-value = 0.038).}
    \label{fig:practice-task}
    \vspace{-0.2cm}
\end{figure}

\noindent \textbf{Policy learning.} We evaluate the performance of policy learning through task success rates. The success rates of the best-performing checkpoint of each learned policy under different modalities is plotted in \cref{fig:success_rate}. We see that kinesthetic teaching leads to the best-performing models for both Open Drawer and Flip Glass, with a large margin in Open Drawer. However, in Push Sanitizer (requiring high contact force), kinesthetic teaching falls short since replaying recorded end-effector poses did not succeed at this task in the first try. The error term we introduced in~\cref{sec:method} induces high jerkiness in actions despite successfully completing the task on the second iteration.

\medskip

\begin{figure}
    \centering
    \includegraphics[width=0.99\linewidth]{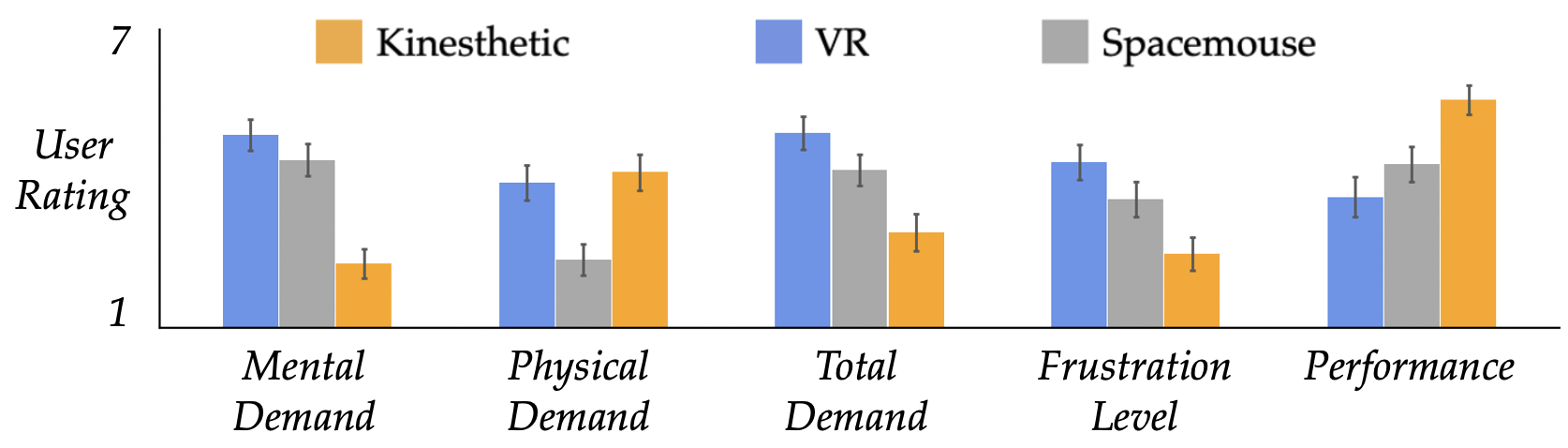}
    \caption{\textbf{User subjective feedback results.} Average 7-point likert scale rating for different demonstration modality is plotted for each question.}
    \label{fig:user_study}
    \vspace{-0.2cm}
\end{figure}

\begin{figure}
    \centering
    \includegraphics[width=0.95\linewidth]{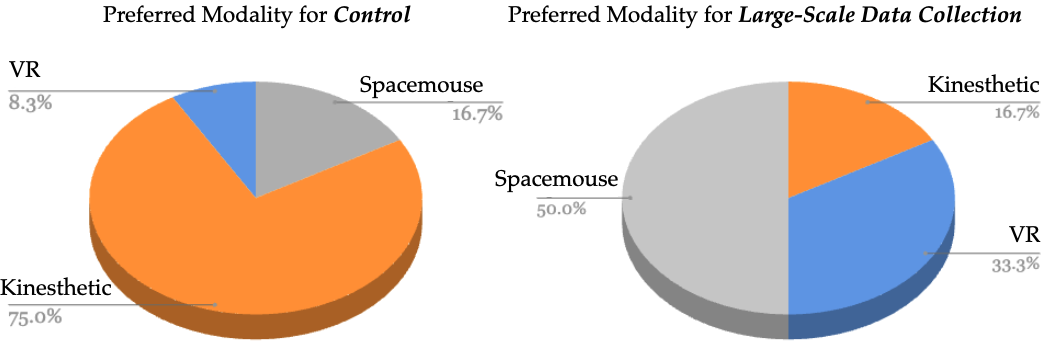}
    \caption{\textbf{Survey question response.} Composition of user responses to questions in final survey. Most participants prefer kinesthetic teaching as the modality for controlling a robot arm to complete a specific task. However, most participants choose teleoperation methods over kinesthetic teaching when asked to perform large-scale data collection.}
    \label{fig:survey}
    \vspace{-0.2cm}
\end{figure}

\noindent \textbf{Intuitiveness (H1).} \cref{fig:practice-task} shows that participants on average spent considerably less time practicing with kinesthetic teaching, supporting \textbf{H1} with a p-value of 0.038 (<0.05). 
This result aligns with prior research showing that users find kinesthetic teaching more intuitive for controlling the robot~\cite{duan2023ar2,jiang2024comprehensive}.
We observe that users find controlling rotation motions with VR unintuitive as they need to constantly reset the reference pose.

\begin{figure*}
    \centering
    \includegraphics[width=\linewidth]{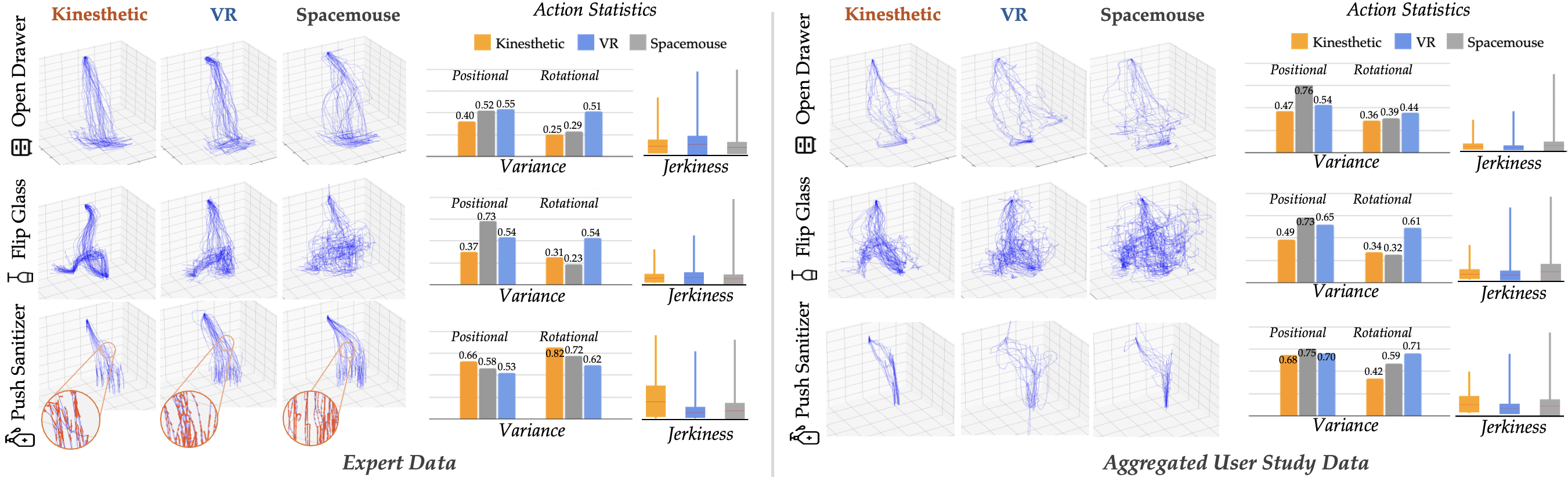}
    \caption{\textbf{Data analysis.} We show visualizations of 3D end-effector positional trajectories and corresponding action statistics across tasks for both (single) expert data and aggregated user study data. We observe that kinesthetic provides data exhibits lower action variance and lower maximum-jerkiness than teleoperation methods, except for the Push Sanitizer task which requires strong contact force. }
    \label{fig:data_analysis}
    \vspace{-0.2cm}
\end{figure*}

\medskip
\noindent \textbf{User experience (H2).} 
The subjective feedback for each modality from 7-scale likert questions is shown in \cref{fig:user_study}. Kinesthetic teaching is consistently rated to require less mental demand ($p\ll0.01$) and frustration level ($p\ll0.01$) and provide highest perceived performance ($p\ll0.01$). 
It is rated to require the highest level of physical demand ($p = 0.016$). 
We do not observe a different trend for the Push Sanitizer task in the relative ratings for kinesthetic teaching, despite the fact that it requires double the amount of time to replay.
However, although users rate kinesthetic teaching to require the least amount of total demand, when asked about which modality they would use for large-scale data collection at the end of the session, the majority of the users choose teleoperation-based modality over kinesthetic teaching (see \cref{fig:survey}). 
This result supports \textbf{H2} and correlates with the status quo that the majority of large-scale datasets are collected through teleoperation-based methods (\cref{fig:modality_percentage}).
Despite the popularity of VR in practical adoption for collecting demonstration data, most participants find it to require high mental demand and lead to high level of frustration in comparison to the other two modalities.

\medskip
\noindent \textbf{Data quality.} We visualize the end-effector trajectories through plotting its position in 3D and compute corresponding action statistics for each task using prescribed data quality metrics from \cref{sec:prelim} (see \cref{fig:data_analysis}).  
Due to high noise and cross-subject variance in non-expert data (6 participants), we discard them for this analysis.
We plot both the data from the single expert demonstrator and aggregated data from user study participants who rated themselves as \emph{Expert} in controlling robots. In general, data collected through teleoperation spreads out in space and covers more diverse states than kinesthetic teaching. 
To estimate action consistency, we compute the average action variance (\cref{eq:av}) for the top $K$ nearest neighbors in proprioceptive state space ($K=200$ for expert data, $K=50$ for user data), and plot positional and rotational action variance separately. We also compute the jerkiness as the second derivative of recorded actions.
We observe that kinesthetic teaching leads to data with relatively low action variance in the two tasks without contact force, but high action variance and jerkiness in the Push Sanitizer task. We further visualize the recorded actions of Push Sanitizer (red arrows inside red circles) and observe that the actions from kinesthetic teaching replay are highly jerky and do not align well with the trajectory.
In the Flip Glass task, the optimal trajectory operates near the joint limits of the robot and therefore the trajectories overlap a lot in kinesthetic teaching mode, especially with data from the single expert demonstrator. However, recruited participants seem to struggle to effectively solve this task through teleoperation modalities, as indicated by the wide spreading trajectories in the plots.

\medskip

\begin{figure}
    \centering
    \includegraphics[width=\linewidth]{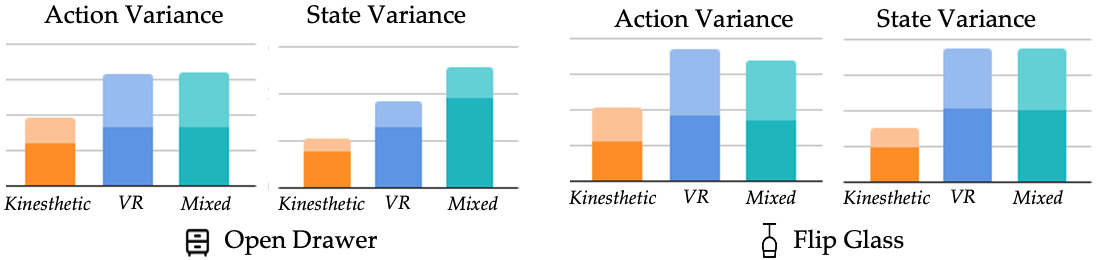}
    \caption{\textbf{Data quality.} Mixing data from kinesthetic and VR can improve state diversity or action consistency of the demonstration dataset (lighter color is rotational variance and darker color is positional variance).}
    \label{fig:mixed-data}
    \vspace{-0.2cm}
\end{figure}

\begin{figure}
    \centering
    \includegraphics[width=0.9\linewidth]{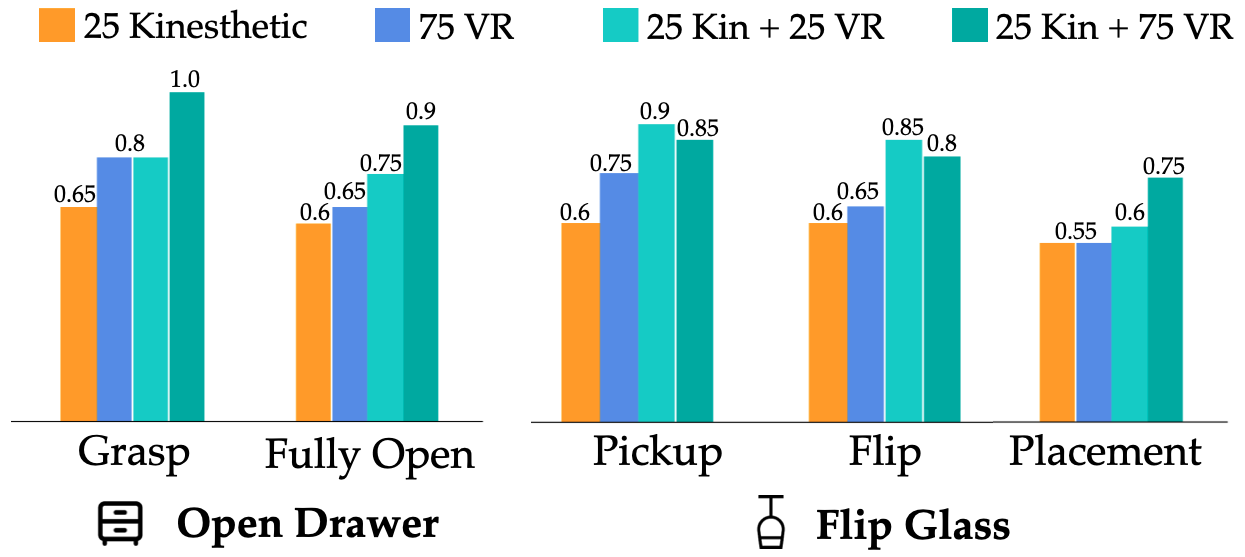}
    \caption{\textbf{Performance of policies using mixed data.} Staged success rates for the best performing models are plotted. Models with mix-modality data (cyan) outperform models learning from single-modality data. }
    \label{fig:mixed}
    \vspace{-0.2cm}
\end{figure}

\noindent \textbf{Learning from mixed modality data.} 
With the insights from our data analysis, we propose a simple hybrid data collection paradigm that mixes data from different demonstration modalities to achieve high quality while maintaining low human effort. Specifically, we can leverage a small amount of kinesthetic data and augment it with larger amounts of teleoperation data. 
We experimented with mixing kinesthetic data with VR data in Open Drawer and Flip Glass tasks. Push Sanitizer is not included since kinesthetic data has low quality due to contact force. \cref{fig:mixed-data} show that the mixed datasets display higher state diversity than datasets from any single modality, and slightly higher action consistency than VR data in Flip Glass case. 
\cref{fig:mixed} shows the success rates of models with mixed data and corresponding baselines using single-modality data. 
We see that the best-performing models with mixed data outperform single-modality baseline by 20\% on average. In Flip Glass, the model trained with a total of 100 mixed demonstrations achieves 75\% success rate, outperforming (+5\%) the best model using all 100 kinesthetic demonstrations (\cref{fig:success_rate}).

\section{Conclusion}
\label{sec:discussion}

In this work, we investigate the impact of demonstration modality on imitation learning, focusing on policy performance, data quality, and user experience. 
Our results indicate that users find kinesthetic teaching more intuitive, and the data collected through kinesthetic teaching leads to better learning performance in tasks that do not involve strong contact forces. 
However, due to the physical effort and time required, users prefer teleoperation modalities for large-scale data collection. 
To address this, we proposed a simple data collection paradigm that combines a small amount of kinesthetic data with additional teleoperation data. We show that this approach yields policies with 20\% higher success rate on average compared to using data from individual modalities alone. An interesting direction for future work is to develop methods for automatically determining the optimal ratio of data to collect from different modalities.



\medskip
\noindent \textbf{Acknowledgement}
This work was supported in part by NSF \#2132847 \& \#2218760, ONR N00014-21-1-2298, AFOSR YIP, Cooperative AI Foundation, and DARPA project \#W911NF2210214. Views and conclusions are of the authors.

\small
\bibliographystyle{IEEEtranN}
\bibliography{reference}

\end{document}